%% file: main.tex
\documentclass[10pt]{article} 
\usepackage[preprint]{tmlr}

\input{math_commands.tex}

\usepackage{hyperref}
\usepackage{url}
\usepackage{graphicx}
\usepackage{subcaption}
\usepackage{booktabs}
\usepackage{multirow}
\usepackage{amsmath}
\usepackage[ruled,vlined]{algorithm2e}

\title{Controllable GNN Explanations via Multi-Metric Preference Selection}

\author{\name Rachit Verma \email \\
      \addr Indian Institute of Technology Gandhinagar
      \AND
      \name Yashraj Deshmukh \email \\
      \addr Indian Institute of Technology Gandhinagar
      \AND
      \name Anirban Dasgupta \email \\
      \addr Indian Institute of Technology Gandhinagar}

\def\month{MM}  
\def\year{YYYY} 
\def\openreview{\url{https://openreview.net/forum?id=XXXX}} 

\begin{document}

\maketitle

\begin{abstract}
Mechanisms for generating GNN explanations are crucial for building trust and mitigating biases in Graph Neural Networks (GNNs), especially in high-stakes scenarios. Most current methods optimize only for fidelity under the sparsity constraint. However, this discounts the need for interpretable explanations (those that consist of familiar motif patterns) and stable explanations (those that remain unchanged under structural perturbations). We propose a novel approach that optimizes GNN explanations across these metrics, exposing their relative weighing as a control. Experiments on various real-world datasets, including MUTAG, BA-2Motif, BAMultiShapes, and PROTEINS, suggest that our method produces higher-fidelity explanations than a state-of-the-art baseline on MUTAG and PROTEINS across all evaluated budgets, and on BA-2Motif at larger budgets, while being faster in the regime of small explanation budgets. They also indicate that our method produces more stable explanations for the MUTAG and BA-2Motif datasets.  We also examine the relationship among different metrics via their induced trade-off surface and explore its dependence on the nature of the motif library.
\end{abstract}

\section{Introduction}

Explainable Machine Learning (XML) has become increasingly crucial in modern machine learning, providing practitioners and regulatory bodies with essential insights into the decision-making mechanisms of complex models\footnote{LLMs were used solely for grammatical editing, text polishing, and generating plotting scripts for figures. All conceptual ideas, methodology, core code implementations, and experimental evaluations presented in this work were developed independently by the authors.}. These interpretability guarantees are particularly important when deploying black-box architectures in high-stakes operational domains. In the context of Graph Neural Networks (GNNs), an explanation $G' \subseteq G_T$ for a target graph $G_T$ is typically formulated as a compact subgraph highlighting the most salient node and edge configurations for the prediction. Within the broader GNN explainability literature, explanation quality is conventionally evaluated across a multidimensional scope comprising fidelity, stability, and interpretability~\citep{yuan2022explainability}. Despite this multi-faceted evaluation paradigm, existing GNN explainer frameworks overwhelmingly optimize for single-objective quantifiers, predominantly prioritizing fidelity~\citep{yuan2022explainability}. However, the operational utility of each metric varies with practitioner requirements and domain-specific constraints. Consequently, there is a need for explainability frameworks that treat subgraph discovery as a multi-objective optimization task, enabling multidimensional customization.

In this work, we focus on three fundamental evaluation metrics of fidelity, interpretability, and stability, each capturing distinct dimensions of explanation quality. Because these metrics exhibit instance and model-dependent correlations, we formulate explanation generation as a multi-objective optimization problem across this metric triad. Operating under the principle that no universally optimal tradeoff exists, our framework explicitly parameterizes relative metric preferences via a practitioner-driven control mechanism (termed the metric-importance controls).



We quantify interpretability based on the presence of human-understandable motifs in the explanation subgraph, consistent with motif-centric evaluation protocols~\citep{yu2023motifexplainermotifbasedgraphneural}. Consistent with stability principles in feature attribution~\citep{alvarezmelis2018robustnessinterpretabilitymethods} and graph explainability~\citep{yuan2022explainability}, stability is defined as the resistance of an explanation to topological perturbations that preserve the underlying graph features. We employ the prevalent formulation of fidelity, as defined by \citet{yuan2022explainability}, which is a combined estimate of how much the explanation's quality drops when the explanation graph is removed, and of the quality of the explanation graph itself.

Formally, a GNN explainer $\mathcal{F}$ is defined as a mapping $\mathcal{F}: \mathcal{M} \times \mathcal{G} \rightarrow \mathcal{G}$, where $\mathcal{G}$ denotes the space of input graphs and $\mathcal{M}: \mathcal{G} \rightarrow \{0,1\}$ represents the space of binary graph classification models. Throughout this work, we refer to the model under evaluation as the target model and the input instance as the target graph. Because defining a closed-form, non-recursive objective for explanation stability is non-trivial, we formulate a two-stage optimization framework to approximate this recursive property. The first stage optimizes for non-recursive metrics, namely the fidelity and interpretability metrics, weighted according to the metric-importance controls. Subsequently, the second stage refines the candidate explanations by additionally optimizing for stability and explicitly enforcing similarity with first-stage explanations generated from perturbed variants of the target graph. To maintain semantic fidelity during this process, we introduce a biased perturbation strategy that preserves the core topological and feature properties of the target graph. We instantiate this strategy with two different graph similarity functions.

Concretely, our contributions in this work include:

\begin{itemize}
    \item Metric Formalization: We formalize the fidelity metric and construct computationally tractable formulations for both graph interpretability and stability metrics.
    \item Controllable Two-Stage Optimization: We propose a two-stage optimization procedure that generates GNN explanations \textit{controllable} by the metric-importance controls.  Empirical evaluations against \textsc{SubgraphX}~\citep{yuan2021explainability} demonstrate that our framework achieves higher fidelity on MUTAG at all budgets and on BA-2Motif at the larger budgets, by margins that exceed the measured seed-to-seed spread of both methods, with the advantage on PROTEINS being only suggestive.

    \item Empirical Trade-off Surface Mapping: We systematically map the multi-metric trade-off surface for both the MUTAG and BA-2Motif datasets, and provide the root cause for the difference in their observations.
    \item Computational Efficiency Analysis: We provide a runtime comparison with \textsc{SubgraphX} and characterize the computational regimes in which our approach offers a speedup.

\end{itemize}

\section{Related Work}

Graph Neural Networks have demonstrated strong performance across relational domains, but the opacity of recursive message passing makes explainability a persistent challenge~\citep {yuan2022explainability}. We group prior explainers by what they search over.

\textbf{Attribution and subgraph search.} Early post-hoc explainers score individual nodes and edges. \textit{GNNExplainer}~\citep{GNNExplainer} maximizes mutual information between the prediction and a continuous edge mask; \textit{PGExplainer}~\citep{luo2020parameterized} amortizes mask generation across a dataset with a learned network; \textit{GISST}~\citep{GISST} combines self-attention with sparsity regularization; \citet{NEURIPS2021_99bcfcd7} adds class-aware pre-training before instance-specific fine-tuning; and removal-based paradigms~\citep{removal_based_attr} connect subgraph sampling to fidelity. Because isolated masks often miss higher-order structure, later work searches over connected subgraphs directly. \textsc{SubgraphX}~\citep{yuan2021explainability}, our baseline, casts this as tree search and uses MCTS to maximize Shapley values. \textit{MotifExplainer}~\citep{yu2023motifexplainermotifbasedgraphneural} instead extracts motifs from a predefined library using attention. Our interpretability metric draws on this motif-centric view, but treats the library as a practitioner-supplied input rather than something generated by heuristics. \textit{MatchExplainer}~\citep{MatchExplainer} matches a target against counterpart graphs by node correspondence, and \textit{GNNXemplar}~\citep{armgaan2025gnnxemplarexemplarsexplanations} grounds importance in dataset-wide exemplars.

\textbf{The evaluation--optimization gap.} Explanations are evaluated on several properties but optimized for one. \textit{Zorro}~\citep{Zorro} recognizes that validity, sparsity, and perturbation robustness matter jointly, and proposes a rate-distortion objective solved by greedy search. However, like its predecessors, it fixes the relative importance of those properties inside a single scalar. Even though our objective is also a scalarization, the difference is that we expose its weights as practitioner-facing controls rather than fixing them, which makes the trade-off among fidelity, interpretability, and stability something the user can inspect and choose a point on. Section~\ref{sec:limitations} discusses what a weighted sum cannot reach, even so.

\section{Methodology}
\textbf{Problem Formulation}

\textbf{Notation.} We write $G = (V, E)$ for a graph, $G_T = (V_T, E_T)$ for the target graph (the input instance to be explained), and $G' \subseteq G_T$ for a candidate explanation, meaning a connected subgraph of $G_T$. We refer to the model being explained as the target model, and write $c_T$ for the class label it assigns to $G_T$. The explanation budget $B$ caps the number of edges in an explanation, so the search is over $\{G' \subseteq G_T : |E(G')| \leq B\}$, where $E(\cdot)$ denotes the edge set of a graph. Perturbed variants of the target graph are written $\tilde{G}$, and the $i$-th sampled perturbation as $\tilde{G}^{(i)}$.

Given the target graph $G_T$, the goal is to find the subgraph $G' \subseteq G_T$ that best explains the target model's prediction. This can be expressed by the following recursive equation:

\begin{equation}
\begin{aligned}
Exp(G_T, B) = \underset{\substack{G' \subseteq G_T, \\ |E(G')| \leq B}}{\arg\max} \Big[ &
w_f \cdot fid(G', G_T) + w_i \cdot interp(G') \\
& + w_s \cdot \mathbb{E}_{\tilde{G} \sim P(\cdot \mid G_T, \theta)} \left[ sim(G', Exp(\tilde{G}, B)) \right] \Big]
\end{aligned}
\label{eq:main_equation}
\end{equation}

The term $fid(G', G_T)$ quantifies the fidelity of an explanation $G'$ with respect to its parent graph, such that $fid: \mathcal{G} \times \mathcal{G} \rightarrow \mathbb{R}$, while $interp(G')$ quantifies its interpretability, where $interp: \mathcal{G} \rightarrow \mathbb{R}$. The function $sim$ is a similarity function such that $sim: \mathcal{G} \times \mathcal{G} \rightarrow \mathbb{R}$. The weights $w_f$, $w_i$, and $w_s$ are the metric-importance controls that specify a point on the tradeoff surface. Throughout this paper, metric-importance controls are written in the order $(w_f, w_i, w_s)$, that is, fidelity first, then interpretability, then stability. We also state weights in normalized form, so that $w_f + w_i + w_s = 1$.

The perturbation distribution is itself defined by the similarity function, $P(\tilde{G} \mid G_T, \theta) = sim_\theta(\tilde{G}, G_T) / \mathcal{Z}$ with $\mathcal{Z}$ a normalizing constant. Here $\theta$ denotes the parameters of the similarity function itself: the encoder weights of the trained VGAE when $sim$ is instantiated by VGAE embeddings, and the kernel's depth and block configuration when it is instantiated by the GNTK (Section~\ref{subsec:similarity}). Consequently, the similarity function $sim(\cdot, \cdot)$ is utilized to define both the overall optimization objective $Exp$ and the parametric probability distribution $P$.

The primary challenge in directly solving this optimization problem is the recursive nature of the stability term, which makes it difficult to derive an analytical expression for the objective. To make the objective tractable, we define a surrogate objective as follows:

\begin{equation}
\begin{aligned}
Exp_s(G_T, B) = \underset{\substack{G' \subseteq G_T, \\ |E(G')| \leq B}}{\arg\max} \Big[ &
w_f \cdot fid(G', G_T) + w_i \cdot interp(G') \\
& + w_s \cdot \mathbb{E}_{\tilde{G} \sim P(\cdot \mid G_T,\theta)} \left[ sim(G', Exp_{it}(\tilde{G}, B)) \right] \Big]
\end{aligned}
\label{eq:surrogate}
\end{equation}

where $Exp_{it}$ is defined as:

\begin{equation}
\begin{aligned}
Exp_{it}(G, B) = \underset{\substack{G' \subseteq G, \\ |E(G')| \leq B}}{\arg\max} \Big[
w_f \cdot fid(G', G) + w_i \cdot interp(G') \Big]
\end{aligned}
\end{equation}

As a result, we approximate Equation~\ref{eq:main_equation} by defining the stability term through the non-recursive, tractable function $Exp_{it}$. We mirror our empirical optimization process to the surrogate objective in Equation~\ref{eq:surrogate} by using a Monte Carlo estimate of the stability term. The first stage produces $N$ perturbed graphs $\{\tilde{G}^{(i)}\}_{i=1}^N$ by sampling them from $P$, along with their corresponding explanations $Exp_{it}(\tilde{G}^{(i)}, B)$. The second stage then optimizes

\begin{equation}
\begin{aligned}
Exp_s(G_T, B) = \underset{\substack{G' \subseteq G_T, \\ |E(G')| \leq B}}{\arg\max} \Big[
& w_f \cdot fid(G', G_T) + w_i \cdot interp(G') \\
& + w_s \cdot stab\big(G', \{\tilde{G}^{(i)}\}_{i=0}^{N}\big) \Big]
\end{aligned}
\label{eq:empirical}
\end{equation}

where $stab$ is the empirical stability estimator defined in Section~\ref{subsec:stability}.

\textbf{Stage 1: Fidelity and Interpretability Optimization.}
\label{subsec:stage1}
Stage 1 finds, for a given graph, the subgraph that maximizes the combined score of fidelity and interpretability weighted by their metric-importance controls $w_f$ and $w_i$. It is run once on the target graph $G_T$ and once on each perturbed graph $\tilde{G}^{(i)}$, so a single explanation of $G_T$ invokes it $N+1$ times in total. The key components are defined as follows.

\textbf{Fidelity Characterization.} We define fidelity through a characterization score~\citep{pytorchgeometric, yuan2022explainability} that combines $fid^+$~\citep{Pope_2019_CVPR}, which measures the \emph{magnitude} of the change in the model's confidence when the explanation is removed, and $fid^-$~\citep{yuan2022explainability}, which measures how well the explanation \emph{alone preserves} that confidence. Let $G' \subseteq G$ be an explanation subgraph of a graph $G$. Given the predicted target class $c$, we denote $p_{\text{orig}}$, $p_{\text{sub}}$, and $p_{\text{comp}}$ as the prediction probabilities for class $c$ on the original graph $G$, the explanation subgraph $G'$, and the complementary graph $G \setminus G'$, respectively. We take


\begin{equation}
fid^+ = \left| p_{\text{orig}} - p_{\text{comp}} \right|, \qquad
fid^- = 1 - \left| p_{\text{orig}} - p_{\text{sub}} \right|,
\end{equation}

so that both components lie in $[0, 1]$ and larger is better in each. The fidelity function is their weighted harmonic mean:

\begin{equation}
fid(G', G) = \frac{w_+ + w_-}{\dfrac{w_+}{fid^+ + \epsilon} + \dfrac{w_-}{fid^- + \epsilon}}
\end{equation}

where $w_+ = w_- = 0.5$ and $\epsilon = 10^{-2}$ prevents division by zero. The harmonic mean penalizes extreme disparities between the two aspects


\textbf{Interpretability Quantification.} We define interpretability as the partial motif-matching score from our subgraph-matching system. For a set of user-defined query graphs (functional groups and chemical motifs in molecular datasets, structural patterns in synthetic graphs) called the \textit{motif library}, we determine the extent to which a query graph $G_q$ is contained in a parent graph $G_p$ using the order-embedding violation penalty of \textsc{NeuroMatch}~\citep{rex2020neuralsubgraphmatching}. Writing $\mathbf{z}_q = \textsc{NeuroMatch}(G_q)$ and $\mathbf{z}_p = \textsc{NeuroMatch}(G_p)$ for the two embedding vectors in $\mathbb{R}^d$,

\begin{equation}
S(G_p, G_q) = \frac{1}{1 + \left\lVert \mathrm{ReLU}\!\left( \mathbf{z}_q - \mathbf{z}_p \right) \right\rVert_2^2},
\label{eq:interp_match}
\end{equation}

with the squared Euclidean norm over the $ d$-dimensional embedding. \textsc{NeuroMatch} is trained so that $\mathbf{z}_q \preceq \mathbf{z}_p$ coordinate-wise when $G_q$ is a subgraph of $G_p$, so the ReLU penalizes only the coordinates that violate containment. $S \in (0, 1]$ therefore provides the MCTS step with a smoother signal than a binary graph-isomorphism reward.

Each match is weighted by a correlation prior $C(G_q)$, and the interpretability of an explanation is
\begin{equation}
interp(G') = \sum_{G_q \in \mathcal{Q}} S(G', G_q) \cdot C(G_q),
\end{equation}
summed over the motif library $\mathcal{Q}$. Note that the parent graph is not an argument: interpretability is defined purely by the motif library and the correlation prior, so the same explanation may yield different interpretability scores for different users.

\textbf{The correlation prior C.} $C$ encodes how much each motif is associated with one class rather than the other, and is a per-dataset input to the method. Concretely, we compute per-class motif statistics $\mathrm{corr}[y][q]$ for each label $y \in \{0, 1\}$ and each motif $q$, as the mean number of occurrences of $q$ per graph of class $y$ across coresets of the datasets, normalized by the size of the motif library. Given a target graph with target label $c_T$, the prior used in the sum above is the signed term:
\begin{equation}
C(G_q) = \mathrm{corr}[c_T][q] - \mathrm{corr}[1-c_T][q].
\end{equation}
Motifs that are more common in the target's own target class contribute positively, and motifs more typical of the opposite target class contribute negatively; consequently, $interp$ is signed rather than non-negative. The motif libraries and the resulting $C$ values for all four datasets are listed in Appendix~\ref{app:motifs}. Note that the statistics are computed over a different subset of the dataset than the evaluation set used in the experimental section, in order to avoid label leakage. 

\textbf{Metric normalization.} Since $fid$ and $interp$ are not natively comparable, we rescale each by a per-dataset constant to ensure that the different metrics lie in similar ranges:
\begin{equation}
\widehat{fid} = \frac{fid}{\sigma_f}, \qquad \widehat{interp} = \frac{interp}{\sigma_i},
\label{eq:normalization}
\end{equation}
The constants used, together with the resulting effective contribution of each term to the reward, are reported in Appendix~\ref{app:hyperparams}. All metric-importance controls quoted in this paper act on the rescaled quantities; all \emph{reported} metric values are on their native scales.

The reward optimized by Stage 1 is then
\begin{equation}
R(G', G) = w_f \cdot \widehat{fid}(G', G) + w_i \cdot \widehat{interp}(G').
\end{equation}

\textbf{Search and budget semantics.} Both stages use MCTS, where a state is a node set inducing a connected subgraph and an action adds one node. Sparsity is enforced as a search constraint rather than an objective penalty.


\textbf{Stage 2: Surrogate Objective Optimization}
The primary goal of stage 2 is to produce explanations that are robust to structural perturbations in the input graph. It can also be understood as a method of `smoothing' the output explanations.

\textbf{Perturbed Graph Sampling}
We now detail the sampling procedure used to draw perturbed graph instances from the target conditional distribution $P(\tilde{G} \mid G_T, \theta) \propto sim_\theta(\tilde{G}, G_T)$. To approximate this distribution, we first generate a candidate pool of $M$ perturbed graphs by random walks over $G_T$. We then filter this candidate pool using the similarity function $sim(\cdot, \cdot)$, retaining the top-$N$ most similar instances to form the set of perturbed target graphs $\{\tilde{G}^{(i)}\}_{i=1}^{N}$. Each retained graph is processed through Stage 1 to yield $Exp_{it}(\tilde{G}^{(i)}, B)$, and these first-stage explanations are used to compute the stability term. $M$, $N$, and the random-walk parameters are in Appendix~\ref{app:hyperparams}.

\textbf{Formulations of the Graph Similarity Function}
\label{subsec:similarity}
To instantiate the similarity function $sim(G_1, G_2)$ between two graphs $G_1 = (V_1, E_1)$ and $G_2 = (V_2, E_2)$, we consider two distinct representation frameworks:

\textbf{Variational Graph Autoencoder (VGAE) Embeddings.} We employ a Variational Graph Autoencoder~\citep{kipf2016variationalgraphautoencoders} to project graph nodes into a shared latent space. Passing $G_1$ and $G_2$ through the encoder yields node-level latent matrices $\mathbf{Z}^{(1)} \in \mathbb{R}^{|V_1| \times d}$ and $\mathbf{Z}^{(2)} \in \mathbb{R}^{|V_2| \times d}$, where $d$ is the latent dimension.

We construct a pairwise node similarity matrix $\mathbf{S} \in \mathbb{R}^{|V_1| \times |V_2|}$, where each entry computes the dot product:
\begin{equation}
\mathbf{S}_{ij} = \mathbf{z}_i^{(1) \top} \mathbf{z}_j^{(2)}.
\end{equation}
A greedy bipartite matching strategy is then applied to $\mathbf{S}$ to extract a set of matched node pairs $\mathcal{M} = \{(u, v) \mid u \in V_1, v \in V_2\}$. The overall graph similarity is the average alignment score across all matched pairs:
\begin{equation}
sim_{\text{VGAE}}(G_1, G_2) = \frac{1}{|\mathcal{M}|} \sum_{(u, v) \in \mathcal{M}} \mathbf{S}_{uv}.
\end{equation}

\textbf{Graph Neural Tangent Kernel (GNTK).} Alternatively, we utilize the Graph Neural Tangent Kernel~\citep{du2019graphneuraltangentkernel}, which quantifies graph similarity in the infinite-width limit of GNNs trained via gradient descent. Unlike node-alignment approaches, GNTK directly captures multi-hop neighborhood structures and node features through a recursive kernel function $\mathcal{K}_{\text{GNTK}}(G_1, G_2)$, and we take $sim_{\text{GNTK}} = \mathcal{K}_{\text{GNTK}}$ directly.


\textbf{Formulation of the Stability Metric}
\label{subsec:stability}
The stability estimator measures how closely the explanation under evaluation resembles the explanations recovered from perturbed copies of the target graph, weighting each perturbed explanation by its own Stage 1 quality. Writing $\tilde{G}^{(0)} := G_T$, so that the target graph's own Stage 1 explanation is included alongside the $N$ perturbations, the estimator is

\begin{equation}
stab\big(G', \{\tilde{G}^{(i)}\}_{i=0}^{N}\big) = \sum_{i=0}^{N} R\big(Exp_{it}(\tilde{G}^{(i)}, B),\, \tilde{G}^{(i)}\big) \cdot sim\big(G',\, Exp_{it}(\tilde{G}^{(i)}, B)\big).
\label{eq:stability}
\end{equation}



The reward weighting also has a structural consequence that we return to in Section~\ref{sec:limitations}: because $R = w_f \widehat{fid} + w_i \widehat{interp}$, the stability estimator is by construction a similarity-weighted composition of the other two metrics, rather than an independently defined property of the explanation.

\textbf{The Complete Pipeline}
Algorithm~\ref{alg:pipeline} in Appendix~\ref{app:algorithm} states the full two-stage procedure for explaining a single target graph, collecting the components above into the order in which they run.

\section{Experiments and Results}
\label{sec:experiments}

\textbf{Experimental Setup.}
The experiments were conducted on four publicly available datasets: MUTAG~\citep{debnath1991structure}, BA-2Motif~\citep{luo2020parameterized}, BAMultiShapes~\citep{azzolin2023global}, and PROTEINS~\citep{borgwardt2005protein}. We use the TUDataset~\citep{morris2020tudatasetcollectionbenchmarkdatasets} repository to instantiate the PROTEINS and BAMultiShapes datasets. All of these datasets are binary graph classification tasks.

All MCTS instances use $20$ simulations and exploration parameter $10$, following~\citet{yuan2021explainability}. The model being explained is a GIN-based graph classifier in every case; its architecture, split, and accuracy are in Appendix~\ref{app:model}.

Because MCTS is stochastic, every number carries run-to-run variation. Appendix~\ref{app:variance} quantifies this noise floor over several seeds on a fixed subsample of each dataset, and we hedge claims against it wherever the effect size is comparable to the spread.

\begin{figure*}[t]
  \centering
  \begin{subfigure}[b]{0.48\linewidth}
    \centering
    \includegraphics[width=\linewidth]{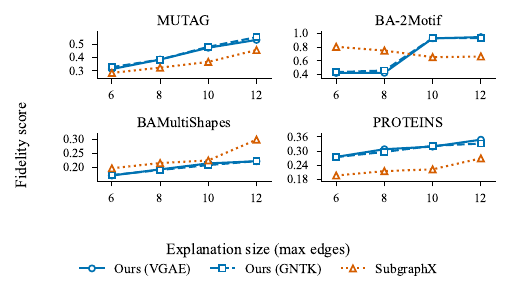}
    \caption{Fidelity}
    \label{fig:sweep-fidelity}
  \end{subfigure}
  \hfill
  \begin{subfigure}[b]{0.48\linewidth}
    \centering
    \includegraphics[width=\linewidth]{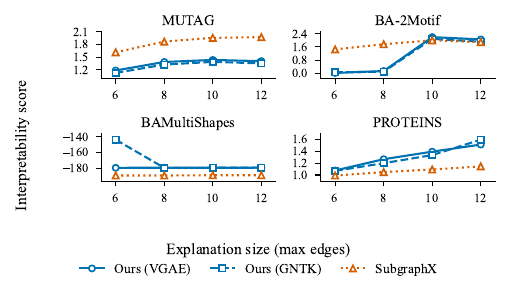}
    \caption{Interpretability}
    \label{fig:sweep-interpretability}
  \end{subfigure}

  \vspace{0.8em} 

  \begin{subfigure}[b]{0.48\linewidth}
    \centering
    \includegraphics[width=\linewidth]{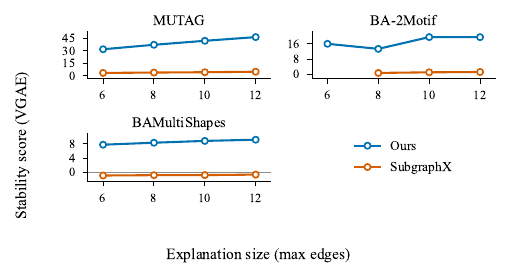}
    \caption{Stability (VGAE)}
    \label{fig:sweep-stability-vgae}
  \end{subfigure}
  \hfill
  \begin{subfigure}[b]{0.48\linewidth}
    \centering
    \includegraphics[width=\linewidth]{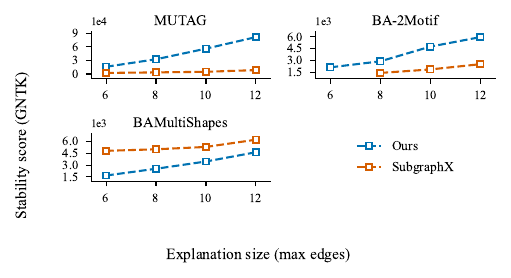}
    \caption{Stability (GNTK)}
    \label{fig:sweep-stability-gntk}
  \end{subfigure}

  \caption{\textbf{Performance sweeps across explanation budgets and datasets.} Our metric-controlled methods against the \textsc{SubgraphX} baseline. \textbf{(a)} Fidelity on native dataset scales. \textbf{(b)} Interpretability. On BA-2Motif, the metric is close to binary, meaning an explanation either fits a planted shape under the budget or it does not, which puts our scores at or near zero at $B \in \{6, 8\}$ . \textbf{(c)} Stability under the VGAE index. \textbf{(d)} Stability under the unbounded GNTK kernel; values are unscaled, and cross-panel comparison with (c) is not meaningful.}

  \label{fig:sweep-all-metrics}
\end{figure*}

\textbf{Comparison with a State-of-the-art Method}

We take \textsc{SubgraphX}~\citep{yuan2021explainability} as the baseline because it uses the same optimization algorithm. We use the DIG~\citep{DIG} implementation with default parameters and $20$ MCTS simulations, matching our setting. Fidelity and interpretability (for both the proposed methods and the baseline) are averaged over the evaluation set (which is simply the complement of the set on which the correlation prior defined earlier is calculated) for each dataset at each budget; stability is averaged over the evaluation set for the proposed method, and the stratified coresets themselves of $200$ graphs ($40$ for MUTAG) for the baseline, due to \textsc{SubgraphX}'s significantly higher runtime. Note that the coresets used to calculate the correlation prior are also used to evaluate SubgrapX's stability results. This does not give rise to a label leakage issue since SubgraphX never sees this label during its optimization process.  The metric-importance controls $(w_f, w_i, w_s)$ are $(0.833, 0.083, 0.083)$, a fidelity-heavy configuration.  We also include a comparison with additional baselines in Appendix~\ref{app:extra_baselines}. 

Scoring \textsc{SubgraphX} on stability requires making use of the biased perturbation strategy introduced earlier; we do so with each index in turn, giving \textsc{SubgraphX-VGAE} and \textsc{SubgraphX-GNTK}, which produce identical explanations to plain \textsc{SubgraphX} and differ only in the stability score assigned (Appendix~\ref{app:subgraphx_variants}). \textsc{SubgraphX} optimizes fidelity alone, so its interpretability and stability columns illustrate the effects of single-objective optimization, rather than a deficiency of the method.

\textbf{Fidelity.} We test each ordering against the combined seed spread of \emph{both} methods, taking a margin as resolved when it exceeds three times $\sqrt{\sigma_{\text{ours}}^2 + \sigma_{\textsc{SubgraphX}}^2}$. On MUTAG, our methods lead at every budget, and the margin resolves at $B \geq 8$ for both indices; at $B = 6$ it resolves under GNTK ($+0.041$ against a threshold of $0.032$) but not under VGAE ($+0.031$ against $0.041$). On BA-2Motif every margin resolves: \textsc{SubgraphX} leads at $B \in \{6, 8\}$ and ours at $B \in \{10, 12\}$, with the crossover between $B = 8$ and $B = 10$. The proposed methods outperform the baseline across all budgets on PROTEINS (with the results being suggestive, as they do not clear the noise floor), whereas the opposite holds on BAMultiShapes (where all the margins resolve).

Our methods exhibit broadly monotone fidelity trends. \textsc{SubgraphX} rises on MUTAG but \emph{declines} monotonically on BA-2Motif, from $0.8$ at $B = 6$ to $0.6$ at $B = 12$. We attribute this to the baseline's top-down pruning mechanism. Given a larger budget, it can retain peripheral edges that dilute the explanation, whereas our method can remove these edges during optimization, even if that means not fully utilizing the explanation budget.


\textbf{Interpretability.}  On MUTAG, the baseline scores are above the proposed methods at all the budgets. On BA-2Motif, the two smallest budgets resolve in \textsc{SubgraphX}'s favor, but the apparent reversal at the two largest does not. The proposed methods outperform the baseline on the PROTEINS dataset across all budgets, though the margins are not resolved for all the budgets, for both indices. The proposed method outperforms the baseline on the BAMultiShapes dataset across all budgets, with all margins clearing the noise floor.

The reversal in trends between fidelity and interpretability for the BAMultiShapes and PROTEINS datasets also points to the nature of the target model. The model scores explanations with higher interpretability as lower fidelity in the case of the former dataset, indicating a clash between the motifs the practitioner wants to obtain in their explanations and the explanations the model flags as high fidelity. In contrast, trends in the latter dataset indicate that it is possible to select explanations with increasing fidelity and interpretability as the budget increases. The proposed method curves for the two metrics on the BA-2Motif dataset exhibit very similar trends, which will be explored further in Section \ref{subsec: tradeoff}.



Datasets of different natures exhibit different results. On MUTAG, whose library holds motifs of varying size, $interp$ rises smoothly until roughly the mean motif size and then yields diminishing returns. For the PROTEINS dataset, the metric continued to rise with increasing budget, indicating that the budget has not yet reached the saturation point. The results of the proposed methods on the synthetic graph datasets show a clear binary behavior in the interpretability metric, with the score increasing abruptly when a certain budget threshold is met. This effect is most pronounced for the BAMultiShapes dataset, due to motifs that are already smaller than the smallest explanation size. Moreover, the \textsc{star} pattern has a significantly higher correlation than the other motifs in the library, leading to its scores dominating the metric computation. We flag this as a limitation of the metric definition.

\cite{yu2023motifexplainermotifbasedgraphneural} is the natural comparison for a motif-based interpretability metric and we do not include it, for want of a reference implementation we could run faithfully.

\textbf{Stability.} \textsc{VGAE} leads \textsc{SubgraphX-VGAE} at every budget on every dataset. Under GNTK, the baseline leads on BAMultiShapes and trails on MUTAG and BA-2Motif. All the mentioned margins clear the noise floor. All methods trend upward with budget, suggesting that larger explanations capture features that are less sensitive to perturbations. Because $stab$ is unbounded and inherits its index's scale, only within-index comparisons are meaningful (Appendix~\ref{app:notes}). These results indicate that our proposed methods can produce more stable explanations while remaining competitive on other metrics. We did not include PROTEINS in the stability comparison due to the high computational cost of evaluating the baseline's stability on this dataset.

Zorro \cite{Zorro} is the natural comparator for a stability claim, being deterministic by construction and explicitly aimed at stable explanations. However, since Zorro optimizes an RDT-fidelity defined over node features, we regard it as inapplicable to the structure-determined graph classification setting we operate in.

\textbf{Input Asymmetry.}
\label{subsec:asymmetry}
Our method receives a motif library and precomputed correlations, while the baseline receives only the target model and graph. In this section, we aim to address this asymmetry by examining the advantages the motif library offers for fidelity metrics.

\textbf{Advantages in the Fidelity Metric.}
We rerun our method with $w_i = 0$, thereby removing the motif library from the objective entirely and isolating any fidelity advantage it provided.


\begin{table}[t]
  \centering
  \caption{Motif-prior ablation on MUTAG at $B=12$, under both similarity
           indices}
  \label{tab:motif-prior-ablation}
  \begin{tabular}{lrrrr}
    \toprule
    & \multicolumn{2}{c}{VGAE} & \multicolumn{2}{c}{GNTK} \\
    \cmidrule(lr){2-3}\cmidrule(lr){4-5}
    Configuration & Fidelity & Stability & Fidelity & Stability \\
    \midrule
    \texttt{interp\_off}         & 0.5433 & 44.32 & 0.5450 & 78\,117 \\
    \texttt{interp\_precomputed} & 0.5449 & 45.69 & 0.5374 & 81\,305 \\
    \bottomrule
  \end{tabular}
\end{table}

\texttt{interp\_off} denotes the proposed method with the metric-importance controls $(w_f, w_i, w_s)$ set to $(0.909, 0, 0.091)$, and \texttt{interp\_precomputed} denotes the configuration of the previous section, $(0.833, 0.083, 0.083)$, which does receive the motif library.

As shown in Table~\ref{tab:motif-prior-ablation}, using the motif library with its precomputed statistics offers almost no advantage in the fidelity and stability metrics. The fidelity difference between the two configurations is $0.0016$ under VGAE and $0.0076$ under GNTK. The seed-to-seed standard deviation of MUTAG fidelity, reported in Appendix~\ref{app:variance}, ranges from $0.001$ to $0.017$ across budgets and indices; at the $B = 12$ configuration used here, it is $0.003$ under VGAE and $0.017$ under GNTK. Both differences are therefore well inside one standard deviation of the run-to-run distribution. We can therefore say that no effect of the motif library on fidelity is detectable at this sample size, not that the effect is zero (Appendix~\ref{app:notes}). Subject to that, a practitioner concerned only with fidelity can use the proposed methods without a motif library at no measurable cost, while retaining the runtime advantage of Section~\ref{sec: runtime}.

\begin{figure*}[t]
  \centering
  \includegraphics[width=\textwidth]{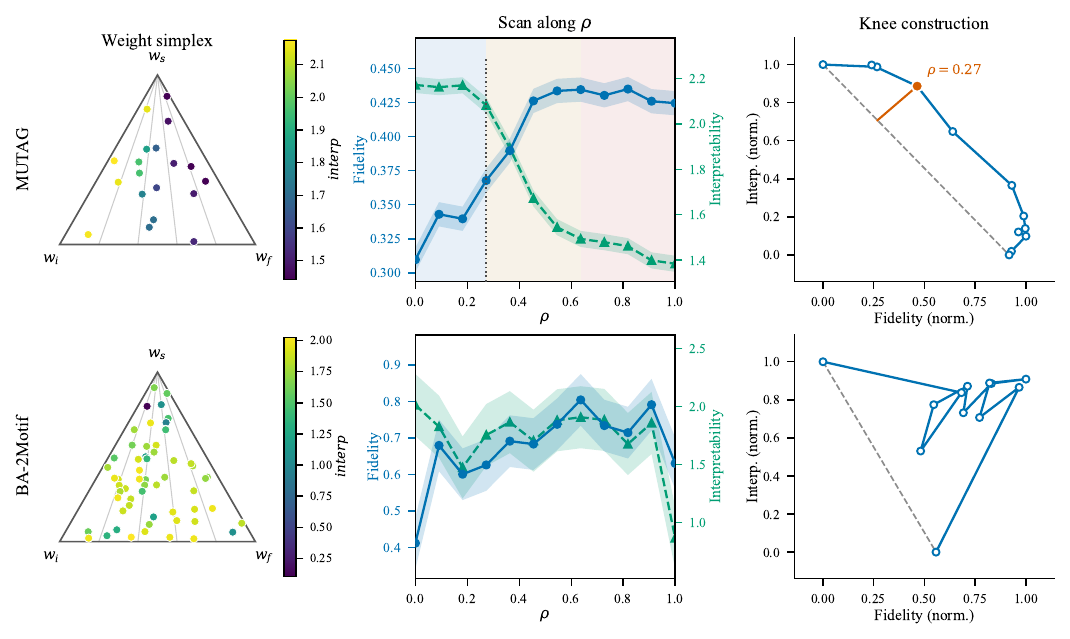}
  \caption{\textbf{The trade-off surface under the VGAE similarity index, for both datasets.}
  Rows are datasets, and the columns are three views of the same sweep.
  \textbf{Left:} sampled weight vectors $(w_i, w_f, w_s)$ on the 2-simplex, coloured by the interpretability of the resulting explanations, with faint rays from the $w_s$ apex marking constant $\rho = w_f/(w_i + w_f)$.
  \textbf{Centre:} scan along $\rho$ at fixed $w_s = \tfrac{1}{3}$, with shaded envelopes giving replicate noise floors.
  \textbf{Right:} knee construction, each metric min-max-normalized.
  The two rows behave qualitatively differently. On MUTAG, fidelity rises, and interpretability falls as $\rho$ increases. On BA-2Motif, the two metrics move \emph{together} so no regime boundaries are identifiable. Section~\ref{subsec:two_regimes} attributes the difference to the composition of each dataset's motif library.}
  \label{fig:tradeoff-vgae}
\end{figure*}

\textbf{The Trade-off Surface.}
\label{subsec: tradeoff}
The central proposal of this paper, that explanation quality is a vector rather than a scalar, raises the question of what the induced trade-off surface actually looks like. We fix $B = 10$ and probe the surface two ways: a Dirichlet sweep of the weight simplex, and a one-dimensional scan along $\rho = w_f/(w_f + w_i)$ at fixed $w_s = \tfrac{1}{3}$. We run both on MUTAG and on BA-2Motif, for both similarity indices. We present the results for the \textsc{VGAE} similarity metric in the main section and those for \textsc{GNTK} in the appendix.






\textbf{Two Trade-off Regimes: Graded and Aligned Motif Libraries.}
\label{subsec:two_regimes}
The scan reveals two qualitatively different surfaces, and which one a dataset exhibits depends on the relationship between its motif library and its labels.

\textbf{Graded libraries give a trade-off.} On MUTAG, fidelity rises, and interpretability falls as $\rho$ increases, and the two are strongly anti-correlated across the scan. Figure~\ref{fig:tradeoff-vgae} (centre) divides this into three regions with boundaries near $\rho = 0.27$ and $\rho = 0.64$: below the first, fidelity rises with interpretability remaining constant; between them, the two exhibit opposing trends; above the second, fidelity has saturated and only interpretability continues to fall. Stability correlates with fidelity, which is consistent with fidelity being the more \emph{consistent} signal for this dataset. Since it is carried by a few specific substructures, perturbing the graph disturbs it less than it disturbs a metric that many subgraphs can satisfy (interpretability).

\textbf{Aligned libraries do not give a trade-off.} On BA-2Motif, the same experiment produces the opposite picture. Fidelity and interpretability are positively correlated, and the simplex does not collapse ($R^2$ between $0.001$ and $0.66$). Consequently, there are no regions to identify based on the value of $\rho$. This also explains why the value of interpretability (which is strongly correlated with fidelity) broadly decreases as $w_s$ increases: since many subgraphs would attain a higher stage 1 score due to the reward signal being less sharp for this dataset, stage 2 produces explanations that are a hybrid of the different stage 1 graphs with increasing $w_s$.

\textbf{Why the two regimes arise.} The difference is a property of the motif library. MUTAG's library contains twenty distinct chemically motivated motifs of varying sizes (Appendix~\ref{app:motifs}), whose label associations are graded and only partially aligned with the classifier's decision boundary. As a result, the two objectives compete. BA-2Motif's library contains exactly the two shapes from which the dataset is constructed, and the planted shape also determines the label, so the most interpretable subgraph \emph{is} the most faithful one, and the objectives are optimized together.

\textbf{Importance of the stability control.} We evaluate linear models predicting explanation outcomes from the ratio $\rho = w_f / (w_f + w_i)$ alone, against a bivariate model incorporating the explicit stability weight $w_s$ (Table~\ref{tab:rho_correlation}). The empirical goodness-of-fit indicates different roles of $w_s$. On MUTAG, $\rho$ alone accounts for the vast majority of metric variance ($R^2 \in [0.678, 0.946]$ across all similarity metric configurations), and introducing $w_s$ yields minimal marginal gain ($\Delta R^2 \le 0.056$, averaging $<0.02$). This indicates that explicit stability control is largely redundant on MUTAG; this is not a surprise, as fidelity and stability are strongly correlated for this dataset. Conversely, on BA-2Motif, $\rho$ alone fails to capture metric behavior, explaining under $35\%$ of fidelity variance and less than $3.3\%$ of interpretability variance. Incorporating $w_s$ provides substantial explanatory recovery, with $\Delta R^2 = +0.199$ for VGAE Fidelity and $+0.135$ for VGAE Interpretability. 





\begin{table}[t]
  \centering
  \caption{Variance in each output metric explained by $\rho = w_f/(w_f + w_i)$ alone, versus by $\rho$ together with $w_s$. Confidence intervals are percentile bootstrap over the $n$ runs per index. The MUTAG results indicate that, although stability control is nearly inert because fidelity is strongly correlated with stability, it still plays a role in determining the metrics for the BA-2Motif dataset, since $\rho$ alone cannot explain the results.}
  \label{tab:rho_correlation}
  
  \begin{tabular}{llcccc}
  \toprule
  \textbf{Dataset} & \textbf{Metric} & \textbf{$R^2$ ($\rho$ only)} & \textbf{95\% CI} & \textbf{$R^2$ ($\rho + w_s$)} & \textbf{Gain from $w_s$} \\
  \midrule
  \multirow{6}{*}{MUTAG ($n{=}20$)} & VGAE Fidelity & 0.765 & [0.592, 0.889] & 0.783 & +0.018 \\
   & VGAE Interpretability & 0.922 & [0.871, 0.962] & 0.930 & +0.007 \\
   & VGAE Stability & 0.678 & [0.430, 0.844] & 0.682 & +0.004 \\
   & GNTK Fidelity & 0.718 & [0.500, 0.865] & 0.773 & +0.056 \\
   & GNTK Interpretability & 0.946 & [0.903, 0.983] & 0.951 & +0.005 \\
   & GNTK Stability & 0.919 & [0.848, 0.972] & 0.945 & +0.026 \\
  \midrule
  \multirow{6}{*}{BA-2Motif ($n{=}60$)} & VGAE Fidelity & 0.259 & [0.087, 0.483] & 0.458 & +0.199 \\
   & VGAE Interpretability & 0.001 & [0.000, 0.073] & 0.135 & +0.135 \\
   & VGAE Stability & 0.646 & [0.554, 0.730] & 0.646 & +0.000 \\
   & GNTK Fidelity & 0.342 & [0.150, 0.563] & 0.444 & +0.102 \\
   & GNTK Interpretability & 0.033 & [0.000, 0.151] & 0.117 & +0.084 \\
   & GNTK Stability & 0.656 & [0.446, 0.832] & 0.763 & +0.106 \\
  \bottomrule
  \end{tabular}
\end{table}

\begin{figure*}[t]
  \centering
  \begin{subfigure}[b]{0.48\linewidth}
    \centering
    \includegraphics[width=\linewidth]{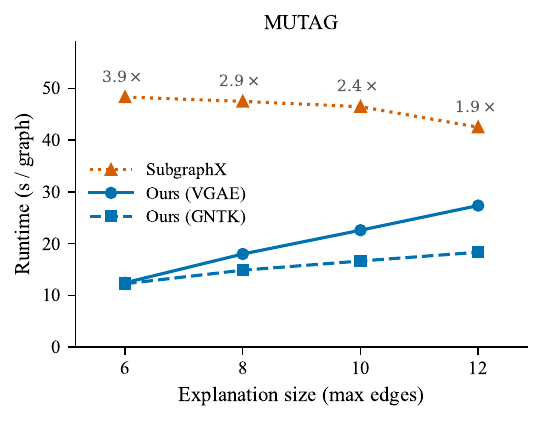}
    \caption{MUTAG}
    \label{fig:runtime-mutag}
  \end{subfigure}
  \hfill
  \begin{subfigure}[b]{0.48\linewidth}
    \centering
    \includegraphics[width=\linewidth]{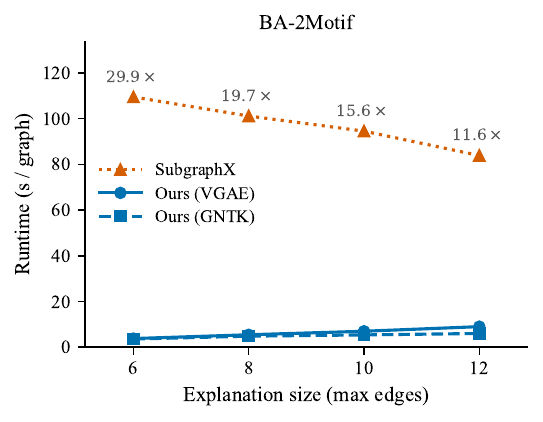}
    \caption{BA-2Motif}
    \label{fig:runtime-ba2motif}
  \end{subfigure}

  \caption{\textbf{Explanation runtime across budgets (seconds per graph).}
  Our method against \textsc{SubgraphX} as a function of explanation size on \textbf{(a)} MUTAG and \textbf{(b)} BA-2Motif. Both are timed at deployment cost. The two frameworks scale in opposite directions.}
  \label{fig:runtime-comparisons}
\end{figure*}

\textbf{Runtime Analysis and Comparison.}
\label{sec: runtime}
 Figures~\ref{fig:runtime-mutag} and \ref{fig:runtime-ba2motif} report mean wall-clock time per graph. Both explainers ran on the same machine with an identical budget of two threads per task and four concurrent tasks, and each figure divides total wall-clock time by the number of graphs completed. Both methods are timed at \emph{deployment} cost, with the stability-scoring work that neither performs in deployment excluded from the timing. Appendix~\ref{app:notes} states the accounting, and Appendix~\ref{app:hyperparams} states the per-phase measurements.



On this basis, our explainer is between $1.9\times$ and $26.7\times$ faster than \textsc{SubgraphX} across the two datasets and all explanation sizes. The two methods exhibit opposing trends as the explanation size increases. \textsc{SubgraphX} becomes more expensive as the target explanation shrinks because its pruning search must descend further to reach a smaller subgraph. Our search grows explanations edge by edge, so a tighter budget means shorter rollouts. The speedup consequently widens as explanations shrink, from $1.9\times$ at $B=12$ to $3.9\times$ at $B=6$ on MUTAG, and from $11.6\times$ to $29.9\times$ on BA-2Motif. This is the regime that matters in practice, since a useful explanation must be substantially sparser than the input graph. \textsc{VGAE} scales more steeply with the budget than \textsc{GNTK}.

\section{Limitations and Future Work}
\label{sec:limitations}

This work introduces a method for computing GNN explanations that is competitive with a state-of-the-art baseline on existing metrics, proposes new ways to evaluate GNN explanations, and illustrates how these metrics interact. Several limitations remain, and they provide directions for follow-up work.

\textbf{Our objective is also a scalarization.} We criticize prior work for collapsing distinct explanation properties into a single scalar objective. However, Equation~\ref{eq:main_equation} is itself a linear scalarization of the same three properties. The difference we claim is not that we avoid scalarizing, but that we \emph{expose} the scalarization: the metric-importance controls are inputs the practitioner sets and can sweep, rather than constants fixed for the method, and Section~\ref{subsec: tradeoff} shows what the resulting surface looks like. This is what allows the explanation quality to be treated as a vector quantity, even though each run returns a single point. Even so, a weighted sum can only recover solutions on the convex hull of the attainable set, so any explanation lying in a concave region of the trade-off surface is unreachable by our search at every weighting. Scalarization schemes that can reach such regions, for instance Chebyshev or $\epsilon$-constraint formulations, would be a natural extension.

\textbf{Better formulations of stability}  A formulation in which a stability term enters the Stage 1 objective directly, so that it can more directly affect the stage 1 search, is the most major single change we can identify. Beyond that, how stability varies with fidelity across models of differing expressivity remains open: in datasets with several discriminative substructures, a stable explanation need not be faithful. Also, restricting the search to graphs on a learned manifold of realistic inputs would prune the space and sidestep some known difficulties in evaluating the true fidelity of subgraphs derived from actual graphs~\citep{zheng2024robustfidelityevaluatingexplainability}.


This work is also meant to serve as a starting point for infusing domain-specific knowledge at various levels of the pipeline. For example, components such as the similarity score, the \textsc{NeuroMatch} encoder, the pruning of the MCTS search, and the tuning of its hyperparameters all require domain-specific knowledge. The work seeks to position GNN explainability as a more involved human-in-the-loop system with greater customization capabilities of the output explanations, so that the practitioner can truly understand the nature of the model-input pair, compared with existing GNN explanation methods that risk being treated as black boxes themselves.


\section{Conclusion}

We address the challenge of incorporating multiple, potentially competing metrics into GNN explanation generation, with a focus on fidelity, interpretability, and stability. We propose formulations for the latter two and a two-stage optimization pipeline that exposes their relative weighting as a user control. We find this method to be comparable to a state-of-the-art method on existing metrics, while being more stable across 2 of the 3 datasets evaluated. We address input asymmetry and show that the proposed method remains competitive even in the absence of user-provided correlations. We also study the induced trade-off surface at a fixed explanation budget. On MUTAG, it collapses to a single fidelity-to-interpretability ratio with three identifiable operating regimes, and the stability control is nearly inert under our surrogate formulation. BA-2Motif exhibits different results because the two metrics are optimized over the same subgraph and therefore never compete, with the stability control having significant weight. Lastly, we discuss the limitations of the present work and point to new avenues for further research.

\bibliography{main}
\bibliographystyle{tmlr}
\appendix

\section{Reproducibility Statement}
\label{app:repro}

All results in this paper were produced by a single code base, submitted as part of the supplementary material. The repository contains the two-stage explainer (\texttt{eval.py}), the reward and metric definitions (\texttt{reward.py}, \texttt{subgraph\_matching.py}), the baseline harness (\texttt{eval\_subgraphx.py}), the motif libraries and correlation statistics as committed JSON files, the and trained target models. Randomness in the \emph{explainer} experiments is controlled by a single \texttt{--seed} flag, present in both our harness and the baseline's, that seeds Python's \texttt{random}, NumPy, and PyTorch; the seed-variance experiment of Appendix~\ref{app:variance} is driven entirely by that flag. All experiments ran on CPU, with two threads per task and four concurrent tasks.

\section{Target Models}
\label{app:model}

\paragraph{Architecture.} Every target model is a three-layer GIN~\citep{xu2019powerfulgraphneuralnetworks} with sum pooling and a single linear readout. Each GIN layer applies a two-layer MLP. The PROTEINS model differs in one respect: its MLPs interleave batch normalisation, as \texttt{Linear--BatchNorm--ReLU--Linear}. The other three use \texttt{Linear--ReLU--Linear}.


\paragraph{Training.} Table~\ref{tab:model_training} gives the full configuration. All four models were trained with a held-out split. We report their accuracy over the full dataset and flag it as such rather than quoting a test figure we cannot reproduce.


\begin{table}[h]
\centering
\caption{Target model training configuration. Accuracy is measured over the full dataset in every case. The last column is the number of graphs the model disagrees with the ground-truth label on, which matters here because the interpretability prior is indexed by the \emph{predicted} class (Appendix~\ref{app:motifs}).}
\label{tab:model_training}
\small
\begin{tabular}{lrrrrrrr}
\toprule
Dataset & Graphs & Split & Seed & Hidden & Optimiser & Epochs & Accuracy (errors) \\
\midrule
MUTAG         &  188 & $80/20$   & ---   & 300 & Adam $3\!\times\!10^{-3}$ & 50  & 0.798 (38) \\
BA-2Motif     & 1000 & $80/20$   & ---   & 300 & Adam $3\!\times\!10^{-3}$ & 50  & 1.000 (0) \\
BAMultiShapes & 1000 & $800/200$ & 12345 &  64 & Adam $1\!\times\!10^{-2}$ & 100 & 0.938 (62) \\
PROTEINS      & 1113 & $890/223$ & 12345 & 128 & Adam $1\!\times\!10^{-3}$ & 100 & 0.845 (172) \\
\bottomrule
\end{tabular}
\end{table}

MUTAG and BA-2Motif were trained with batch size $32$ under cross-entropy; BAMultiShapes with batch size $64$ under a negative log-likelihood applied to the log of the model's softmax output; PROTEINS with batch size $64$, weight decay $10^{-4}$, and cross-entropy on raw logits.

We make no state-of-the-art claim for these models. They are the fixed classifiers both explainers are asked to explain, and no comparison in this paper depends on their absolute accuracy.



\section{Auxiliary Components}
\label{app:auxiliary}

The pipeline uses three learned components besides the target model: a VGAE and the GNTK, which define the two similarity indices, and \textsc{NeuroMatch}, which provides the soft subgraph-containment score behind the interpretability term. This appendix gives their training setup and states two properties of it that determine how the results should be interpreted.

\paragraph{VGAE.} A graph convolutional variational autoencoder with hidden dimension $64$ and latent dimension $32$, trained once per dataset with Adam at learning rate $10^{-2}$ for $10$ epochs under the standard variational reconstruction objective, with \texttt{torch.manual\_seed(12345)} fixed. Training iterates over the graphs of the dataset one at a time; within each graph the edges are split $85/5/10$ into training, validation and test, and the model is trained to reconstruct held-out edges. Similarity between two explanation subgraphs is then computed by greedy node matching in the latent space.

\paragraph{\textsc{NeuroMatch}.} An order-embedding encoder~\citep{rex2020neuralsubgraphmatching} of three GIN layers with hidden dimension $64$, mean pooling, and a two-layer post-pooling MLP producing a $64$-dimensional embedding. A final ReLU constrains embeddings to the non-negative orthant, which is what makes the order relation well defined. It is trained with the max-margin order-embedding loss at margin $1$: for a pair $(M, G)$ with embeddings $z_M, z_G$, the energy $E = \lVert \mathrm{relu}(z_M - z_G) \rVert_2^2$ tends to zero when $M$ is a subgraph of $G$ and above the margin when it is not. Training pairs are drawn by sampling a query motif and a dataset graph uniformly at random and labelling the pair by \emph{exact} subgraph isomorphism. We use $1000$ pairs per epoch, batch size $16$, Adam at learning rate $10^{-3}$, for $20$ epochs.

\paragraph{GNTK.} The graph neural tangent kernel of~\citet{du2019graphneuraltangentkernel}, used as a fixed kernel with no fitted parameters; its structure is given in Appendix~\ref{app:hyperparams}.

\paragraph{Neither auxiliary component sees a class label.} We state this explicitly because the interpretability prior does use labels and a reader may reasonably ask whether label information enters elsewhere. The VGAE objective is edge reconstruction, which is unsupervised. \textsc{NeuroMatch}'s targets are subgraph-isomorphism outcomes, computed combinatorially from the pair of graphs. Neither can leak class information into a similarity score or a containment score.


\paragraph{Reproducibility.} The released checkpoints are provided for reproducing the numbers in this paper.

\section{Experimental Parameters}
\label{app:hyperparams}


\paragraph{Search.} Both stages use MCTS with $20$ simulations per move, exploration constant $C = 10$, and rollout depth $100$. A state is a set of nodes inducing a connected subgraph. An action adds one node. The search expands only to states whose induced subgraph has at most $B$ edges.

\paragraph{Perturbation sampling.} For each target graph we draw a candidate pool of $M = 25$ perturbed graphs by random walks over the target, where each walk is seeded with $\lfloor k \cdot |V_T| \rfloor$ edges and $k$ is drawn uniformly from $10$ values evenly spaced in $[0.6, 0.8]$. The $N = 10$ most similar graphs are then selected as the final perturbation graphs. Stage 1, therefore, runs $N + 1 = 11$ times per target graph: once on the target and once on each retained perturbation.



\paragraph{Metric normalization.} As described in Section~\ref{subsec:stage1}, $fid$ and $interp$ are each divided by a per-dataset constant. We use $\sigma_f = 0.1$ for every dataset, and $\sigma_i = 1000$ for BAMultiShapes and $\sigma_i = 1$ elsewhere. Stability is likewise divided by $1000$ under the GNTK index and left unscaled under VGAE. The constants are round powers of ten chosen from the raw metric means in columns 2-3 of Table~\ref{tab:effective_weights}: fidelity sits near $0.1$--$0.7$ across datasets, and interpretability near unity everywhere except BAMultiShapes, where the star motif's occurrence count (Appendix~\ref{app:motifs}) pushes it into the hundreds. 


After scaling, the two metrics lie within a factor of $2.2$--$8.4$ of each other (Table \ref{tab:effective_weights}), so neither term can dominate the reward by orders of magnitude solely through its units. The \emph{effective} contributions at the default configuration are further apart than the weights alone imply: $3.56$ against $0.146$ on MUTAG, a ratio of $24{:}1$, and $5.88$ against $0.096$ on BA-2Motif, a ratio of $61{:}1$, where the nominal preference is only $10{:}1$. The default configuration is, therefore, in effect, more fidelity-heavy than its weights suggest.

\paragraph{Structure of the GNTK.} Because the GNTK carries no learned parameters, its structure is its specification. Let $G_1, G_2$ have adjacency matrices $A_1, A_2$, degrees $d_u$, and node features $h_u$. The kernel is built from $L$ \emph{blocks}, each consisting of one neighbourhood aggregation followed by $R$ fully-connected ReLU layers. We use $L = 4$, $R = 2$, degree scaling, and no jumping knowledge, which are the defaults of our implementation and are never overridden.

Write $\Sigma$ for the covariance between the two graphs' node representations and $\Theta$ for the accumulated tangent kernel, both matrices over pairs $(u, v) \in V_1 \times V_2$. The recursion is initialised with the feature inner product,
\begin{equation}
\Sigma_{uv} \leftarrow h_u^\top h_v ,
\end{equation}
and each aggregation step sums over neighbourhoods,
\begin{equation}
\Sigma_{uv} \leftarrow c_{uv} \sum_{u' \in \mathcal{N}(u)} \sum_{v' \in \mathcal{N}(v)} \Sigma_{u'v'},
\qquad c_{uv} = \frac{1}{d_u \, d_v},
\label{eq:gntk_agg}
\end{equation}
which is implemented as a Kronecker product $A_1 \otimes A_2$. Each of the $R$ fully-connected layers then applies the arc-cosine recursion for ReLU activations. Writing $\bar{\Sigma}_{uv} = \Sigma_{uv} / \sqrt{\Sigma_{uu} \Sigma_{vv}}$ for the correlation and clipping it to $[-1, 1]$,
\begin{equation}
\begin{aligned}
\Sigma_{uv} &\leftarrow \frac{\sqrt{\Sigma_{uu}\Sigma_{vv}}}{\pi}\Big[ \bar{\Sigma}_{uv}\big(\pi - \arccos \bar{\Sigma}_{uv}\big) + \sqrt{1 - \bar{\Sigma}_{uv}^{2}} \Big], \\
\dot{\Sigma}_{uv} &\leftarrow \frac{\pi - \arccos \bar{\Sigma}_{uv}}{\pi},
\qquad
\Theta_{uv} \leftarrow \Theta_{uv}\,\dot{\Sigma}_{uv} + \Sigma_{uv}.
\end{aligned}
\end{equation}
Aggregation is applied to both $\Sigma$ and $\Theta$ between consecutive blocks but not after the last. The graph-level kernel is the sum of the final block's tangent kernel over all node pairs,
\begin{equation}
\mathcal{K}_{\text{GNTK}}(G_1, G_2) = 2 \sum_{u \in V_1} \sum_{v \in V_2} \Theta_{uv},
\label{eq:gntk_readout}
\end{equation}
With jumping knowledge disabled, intermediate blocks do not contribute to the readout.

Note that nothing here is trained, so the $\theta$ that parameterizes the perturbation distribution in Section~\ref{subsec:similarity} is the tuple $(L, R, \text{scale}, \text{jk})$ rather than a weight vector. Also, the reference implementation adds self-loops to the adjacency matrix before aggregating, whereas ours uses the plain adjacency matrix. With $L = 4$ blocks, the receptive field is unaffected, but the kernel is not numerically identical to the reference.

\begin{table}[h]
\centering
\caption{Effect of the per-dataset normalization, averaged over explanation budgets for the \textsc{VGAE} method, under the reference-corpus prior. Columns 2--3 give the raw metric means, columns 4--5 the same quantities after division by $\sigma_f$ and $\sigma_i$, and columns 6--7 the effective contribution $w \cdot \widehat{\text{metric}}$ to the reward under the default controls $(0.833, 0.083, 0.083)$.}
\label{tab:effective_weights}
\small
\begin{tabular}{lrrrrrr}
\toprule
& \multicolumn{2}{c}{Raw} & \multicolumn{2}{c}{Scaled} & \multicolumn{2}{c}{Effective} \\
\cmidrule(lr){2-3}\cmidrule(lr){4-5}\cmidrule(lr){6-7}
Dataset & $fid$ & $interp$ & $\widehat{fid}$ & $\widehat{interp}$ & $w_f\widehat{fid}$ & $w_i\widehat{interp}$ \\
\midrule
MUTAG         & 0.425 &    1.354 & 4.25 &  1.354 & 3.54 & $+0.112$ \\
BA-2Motif     & 0.681 &    1.103 & 6.81 &  1.103 & 5.67 & $+0.092$ \\
BAMultiShapes & 0.201 & $-179.379$ & 2.01 & $-0.179$ & 1.67 & $-0.015$ \\
PROTEINS      & 0.312 &    1.312 & 3.12 &  1.312 & 2.60 & $+0.109$ \\
\bottomrule
\end{tabular}
\end{table}

\paragraph{Runtime accounting.} The runtimes reported in Section~\ref{sec: runtime} for our method include all $N+1$ Stage 1 invocations per target graph, the perturbation sampling, and the Stage 2 search. 

They exclude one thing: the block that re-samples perturbations after Stage 2 purely to compute the stability score we report, which a deployed explainer would not run. To quantify it we instrumented the pipeline with per-phase timers and ran all eight index--budget cells on MUTAG over $20$ graphs each. Table~\ref{tab:phases} gives the resulting breakdown. The deployment fraction is $0.597 \pm 0.017$ across those eight cells (range $0.579$--$0.620$). Repeating the exercise on BA-2Motif gives $0.672 \pm 0.018$ (range $0.647$--$0.701$). We scale each dataset by its own measured fraction. 



\begin{table}[h]
\centering
\caption{Per-phase cost of explaining one MUTAG graph, averaged over both similarity indices and all four budgets at $20$ graphs per cell. Only \texttt{stab\_eval} is excluded from the reported runtimes.}
\label{tab:phases}
\begin{tabular}{lrl}
\toprule
Phase & s/graph & Role \\
\midrule
\texttt{setup}           &  0.003 & model forward passes, normalization constants \\
\texttt{stage1\_main}     &  1.108 & Stage 1 search on the target graph \\
\texttt{smoothing}       & 10.642 & sampling $M$ perturbations, Stage 1 on the retained $N$ \\
\texttt{stability\_norm}  &  0.001 & stability normalization constants \\
\texttt{stage2\_main}     &  4.260 & Stage 2 search on the target graph \\
\midrule
\texttt{stab\_eval}       & 10.679 & \emph{evaluation only}: re-samples to score stability \\
\bottomrule
\end{tabular}
\end{table}

\section{Motif Libraries and Correlation Priors}
\label{app:motifs}

The motif library $\mathcal{Q}$ is an input to our method, supplied per dataset. The molecular datasets consist of chemically meaningful functional groups and ring systems; the synthetic datasets consist of the planted shapes used to construct them. Proteins consist of secondary-structure arrangements.


$C$ is estimated on a reference corpus held out from every evaluated graph, using the corpus graphs' true labels: $\mathrm{corr}[y][q]$ is the mean number of occurrences of motif $q$ per corpus graph of class $y$, normalized by the library size. Using labels here is the point, since the prior stands in for knowledge a practitioner already holds about which motifs go with which class, and a practitioner's prior is built from labelled data.

$C$ is applied by indexing it with the target model's \emph{predicted} class $c_T$, never the target graph's ground-truth label: the weight on motif $q$ is $\mathrm{corr}[c_T][q] - \mathrm{corr}[1-c_T][q]$. The interpretability term therefore asks whether the class the model chose looks, in motif terms, like what this practitioner would expect for that class. 


Together these give the guarantee that nothing about an evaluated graph's own label enters its own explanation: it is absent from the corpus, and the index is a prediction.


The tables below list both class statistics and the resulting contrast for $c = 1$; the contrast for $c = 0$ is its negation. Each header also carries the cosine similarity between the corpus prior and the prior estimated from the whole dataset. It is $\geq 0.999$ on MUTAG, BA-2Motif and BAMultiShapes, and On PROTEINS it is $0.772 \pm 0.244$, with $61\%$ of corpora below $0.9$.

\paragraph{Counting convention} A motif's per-class statistic is the mean number of \emph{subgraph isomorphisms} per graph of that class, normalized by the library size. A $k$-node star therefore embeds once for every ordered choice of its leaves, so its count grows combinatorially with node degree: a degree-$d$ node contains $\binom{d}{k-1}$ copies of a $k$-star.

This means that on BAMultiShapes the \texttt{star} statistic reaches $6098.22$ where the library's other two motifs stay below $2.7$, so the contrast $-3237.94$ dominates everything else.

\paragraph{Three motif pairs are duplicates.} Testing the libraries for isomorphism under the node labels the matcher actually sees shows \texttt{ethyl} $\equiv$ \texttt{ethene} and \texttt{benzene\_ring} $\equiv$ \texttt{cyclic\_butyl} in MUTAG, and \texttt{beta\_hairpin} $\equiv$ \texttt{core\_clique\_helix} in PROTEINS; each pair have the identical statistics in the tables below. The collisions are a matcher limitation. MUTAG node features encode only atom type and the matcher ignores edge attributes, so bond order is invisible and C--C and C=C are labelled alike. The effect is that the duplicated motif is counted twice, which doubles its weight in $interp$ relative to a library with the collision removed.


\paragraph{$interp$ is signed.} Because $C$ is a contrast between classes, motifs typical of the opposite class enter with negative weight, and $interp$ can be negative. 



\input{motif_tables.tex}

\section{The Two-Stage Pipeline}
\label{app:algorithm}

\begin{algorithm}[h]
\SetAlgoLined
\KwIn{target graph $G_T$, target model $f$, budget $B$, controls $(w_f, w_i, w_s)$, motif library $\mathcal{Q}$, correlation prior $C$, similarity $sim_\theta$, pool size $M$, retained perturbations $N$}
\KwOut{explanation $G^\star \subseteq G_T$}
\BlankLine
\tcp{Stage 1 on the target graph}
$G^{(0)} \leftarrow \textsc{MCTS}\big(G_T,\ R(\cdot, G_T)\big)$ subject to $|E(\cdot)| \leq B$\;
$\mathcal{A} \leftarrow \big\{ \big(G^{(0)},\ R(G^{(0)}, G_T)\big) \big\}$ \tcp*{smoothing set}
\BlankLine
\tcp{Perturbation sampling}
draw a candidate pool $\{\tilde{G}_c\}_{c=1}^{M}$ by random walks over $G_T$\;
$\{\tilde{G}^{(i)}\}_{i=1}^{N} \leftarrow$ the $N$ candidates maximizing $sim_\theta(\tilde{G}_c, G_T)$\;
\BlankLine
\tcp{Stage 1 on each perturbation}
\For{$i \leftarrow 1$ \KwTo $N$}{
  $G^{(i)} \leftarrow \textsc{MCTS}\big(\tilde{G}^{(i)},\ R(\cdot, \tilde{G}^{(i)})\big)$ subject to $|E(\cdot)| \leq B$\;
  \lIf{$G^{(i)} \neq \emptyset$}{$\mathcal{A} \leftarrow \mathcal{A} \cup \big\{ \big(G^{(i)},\ R(G^{(i)}, \tilde{G}^{(i)})\big) \big\}$}
}
\BlankLine
\tcp{Stage 2 on the target graph}
$G^\star \leftarrow \textsc{MCTS}\big(G_T,\ R(\cdot, G_T) + w_s \cdot stab(\cdot, \mathcal{A})\big)$ subject to $|E(\cdot)| \leq B$\;
\Return $G^\star$\;
\caption{Two-stage controllable explanation search}
\label{alg:pipeline}
\end{algorithm}

\section{The \textsc{SubgraphX-VGAE} and \textsc{SubgraphX-GNTK} Variants}
\label{app:subgraphx_variants}

To score the baseline on stability, we must supply the biased perturbation mechanism over which the stability metric is defined. Algorithm~\ref{alg:sgx_variant} presents a definition of the variants. The \textsc{SubgraphX} search itself is unmodified and is re-run on each perturbed graph exactly as it is on the target, and only the perturbation sampler and the similarity function are additionally introduced. Consequently, \textsc{SubgraphX-VGAE} and \textsc{SubgraphX-GNTK} produce \emph{identical} explanations to plain \textsc{SubgraphX} on the target graph, and differ only in the stability score assigned to those explanations. This is why the fidelity and interpretability panels of Figure~\ref{fig:sweep-all-metrics} show a single \textsc{SubgraphX} curve while the stability panels show two.

\begin{algorithm}[h]
\SetAlgoLined
\KwIn{target graph $G_T$, target model $f$, budget $B$, similarity $sim_\theta$, pool size $M$, retained perturbations $N$}
\KwOut{explanation $G^\star$ and its stability score}
$G^\star \leftarrow \textsc{SubgraphX}(G_T, f, B)$ \tcp*{unmodified baseline}
draw a candidate pool $\{\tilde{G}_c\}_{c=1}^{M}$ by random walks over $G_T$\;
$\{\tilde{G}^{(i)}\}_{i=1}^{N} \leftarrow$ the $N$ candidates maximizing $sim_\theta(\tilde{G}_c, G_T)$\;
\For{$i \leftarrow 1$ \KwTo $N$}{
  $G^{(i)} \leftarrow \textsc{SubgraphX}(\tilde{G}^{(i)}, f, B)$ \tcp*{re-run on each perturbation}
}
\Return $G^\star$, $stab\big(G^\star, \{\tilde{G}^{(i)}\}\big)$ using Equation~\ref{eq:stability}\;
\caption{Stability scoring for the \textsc{SubgraphX} baseline}
\label{alg:sgx_variant}
\end{algorithm}

\section{Additional Notes on the Experimental Results}
\label{app:notes}



\paragraph{Properties of the stability estimator.} The stability estimator is a \emph{reward-weighted} sum rather than a plain average: each perturbed explanation contributes in proportion to the Stage 1
reward it achieved, so a perturbation whose explanation was itself poor contributes little. As a direct consequence, $stab$ inherits the scale of both $R$ and $sim$ and is not bounded in $[0, 1]$. Stability values are therefore meaningful only \emph{within} a single similarity index, and we compare methods only within an index throughout.


\paragraph{Reading the motif-prior ablation.} The comparison in Section~\ref{subsec:asymmetry} is a null result and needs to be read against the noise floor. We can say that no effect of the motif library on fidelity is detectable at this sample size, not that the effect is zero.

\section{Additional Baselines: Full Sweep}
\label{app:extra_baselines}

Table~\ref{tab:extra-baselines-full} reports \textsc{GNNExplainer} and \textsc{PGExplainer} on every metric and every budget. Both were run through the same setup as \textsc{SubgraphX}, against the same target models, perturbation sampler, and stability estimator; the only difference is the explainer. \textsc{GNNExplainer} optimizes a per-instance edge mask and \textsc{PGExplainer} amortizes the mask over a trained network, and in both cases, we impose the explanation budget by taking the top-$B$ edges of the resulting mask.

These results need to be read with a caveat: a top-$B$ edge selection need not induce a connected subgraph, whereas both \textsc{SubgraphX} and our method search under a connectivity constraint. This depresses all three metrics for the mask-based methods, with interpretability most severely affected, since a disconnected edge set cannot contain a motif, and stability next, since it is measured as similarity to explanations of perturbed graphs, and disconnected selections agree with each other less. The comparison is therefore informative about how these explainers behave under a hard sparsity budget using these metrics, rather than a general ranking of explanation quality.

\begin{table}[h]
\centering
\small
\caption{\textsc{GNNExplainer} and \textsc{PGExplainer} across all budgets and metrics.}
\label{tab:extra-baselines-full}
\input{baseline_table_full.tex}
\end{table}

\section{Seed Variance}
\label{app:variance}

\input{variance_table.tex}

\clearpage

\section{Trade-off Surface for the GNTK metric}
\label{app:ba2motif_tradeoff}

We present the same figure as \ref{fig:tradeoff-vgae} for the GNTK metric in this section.

\input{ba2motif_tradeoff.tex}

\end{document}

%% file: math_commands.tex
\usepackage{amsmath,amsfonts,bm}

\def\eqref#1{equation~\ref{#1}}

\def\1{\bm{1}}

\DeclareMathAlphabet{\mathsfit}{\encodingdefault}{\sfdefault}{m}{sl}
\SetMathAlphabet{\mathsfit}{bold}{\encodingdefault}{\sfdefault}{bx}{n}



%% file: motif_tables.tex
\subsection*{MUTAG ($|\mathcal{Q}| = 22$, corpus $= 40$ graphs, $\cos = 0.999 \pm 0.001$)}
\begin{tabular}{lrrr}
\toprule
Motif & $\mathrm{corr}[0]$ & $\mathrm{corr}[1]$ & $C$ for $c{=}1$ \\
\midrule
\texttt{amino\_benzene} & 0.1503 & 0.1768 & +0.0264 \\
\texttt{anthracene} & 0.0000 & 0.0404 & +0.0404 \\
\texttt{aromatic\_amine} & 0.0210 & 0.0168 & -0.0041 \\
\texttt{aromatic\_oxy} & 0.0909 & 0.0320 & -0.0589 \\
\texttt{benzene\_ring} & 0.7552 & 1.3131 & +0.5579 \\
\texttt{bromo} & 0.0000 & 0.0000 & +0.0000 \\
\texttt{chloro} & 0.0280 & 0.0017 & -0.0263 \\
\texttt{cyanide} & 0.0874 & 0.0926 & +0.0052 \\
\texttt{cyclic\_butyl} & 0.7552 & 1.3131 & +0.5579 \\
\texttt{dinitro} & 0.0140 & 0.0067 & -0.0073 \\
\texttt{ester\_group} & 0.0000 & 0.0034 & +0.0034 \\
\texttt{ethene} & 0.8322 & 1.3535 & +0.5214 \\
\texttt{ether} & 0.0559 & 0.0135 & -0.0425 \\
\texttt{ethyl} & 0.8322 & 1.3535 & +0.5214 \\
\texttt{fluoro} & 0.0000 & 0.0017 & +0.0017 \\
\texttt{imidazole} & 0.0000 & 0.0000 & +0.0000 \\
\texttt{iodo} & 0.0000 & 0.0017 & +0.0017 \\
\texttt{ketone} & 0.0839 & 0.0236 & -0.0603 \\
\texttt{napthalene} & 0.0000 & 0.2290 & +0.2290 \\
\texttt{nitro\_group} & 0.0979 & 0.1380 & +0.0401 \\
\texttt{propyl} & 0.9231 & 1.8418 & +0.9187 \\
\texttt{pyridine} & 0.0070 & 0.0000 & -0.0070 \\
\bottomrule
\end{tabular}

\subsection*{BA-2Motif ($|\mathcal{Q}| = 2$, corpus $= 200$ graphs, $\cos = 1.000 \pm 0.000$)}
\begin{tabular}{lrrr}
\toprule
Motif & $\mathrm{corr}[0]$ & $\mathrm{corr}[1]$ & $C$ for $c{=}1$ \\
\midrule
\texttt{cycle} & 0.0000 & 0.9300 & +0.9300 \\
\texttt{house} & 4.7000 & 0.0000 & -4.7000 \\
\bottomrule
\end{tabular}

\subsection*{BAMultiShapes ($|\mathcal{Q}| = 3$, corpus $= 200$ graphs, $\cos = 1.000 \pm 0.000$)}
\begin{tabular}{lrrr}
\toprule
Motif & $\mathrm{corr}[0]$ & $\mathrm{corr}[1]$ & $C$ for $c{=}1$ \\
\midrule
\texttt{house} & 0.0990 & 0.4040 & +0.3050 \\
\texttt{star} & 6098.2178 & 2860.2828 & -3237.9350 \\
\texttt{wheel} & 0.5281 & 2.6263 & +2.0982 \\
\bottomrule
\end{tabular}

\subsection*{PROTEINS ($|\mathcal{Q}| = 13$, corpus $= 200$ graphs, $\cos = 0.772 \pm 0.244$)}
\begin{tabular}{lrrr}
\toprule
Motif & $\mathrm{corr}[0]$ & $\mathrm{corr}[1]$ & $C$ for $c{=}1$ \\
\midrule
\texttt{bab\_motif} & 0.8989 & 0.3764 & -0.5225 \\
\texttt{beta\_hairpin} & 3.0769 & 1.6498 & -1.4272 \\
\texttt{core\_clique\_helix} & 3.0769 & 1.6498 & -1.4272 \\
\texttt{core\_clique\_sheet} & 2.4383 & 2.3470 & -0.0913 \\
\texttt{helix\_bundle} & 0.8786 & 0.4081 & -0.4705 \\
\texttt{hub\_node\_1} & 0.1045 & 0.0458 & -0.0587 \\
\texttt{hub\_node\_2} & 0.0000 & 1.5712 & +1.5712 \\
\texttt{hub\_node\_3} & 0.0000 & 0.0022 & +0.0022 \\
\texttt{hub\_node\_4} & 0.0135 & 0.0436 & +0.0301 \\
\texttt{hub\_node\_5} & 0.0116 & 0.0065 & -0.0051 \\
\texttt{hub\_node\_6} & 0.4093 & 0.2193 & -0.1900 \\
\texttt{hub\_node\_7} & 0.0019 & 0.0011 & -0.0008 \\
\texttt{hub\_node\_8} & 0.0058 & 0.0000 & -0.0058 \\
\bottomrule
\end{tabular}

%% file: baseline_table_full.tex
\begin{tabular}{llcccc}
\toprule
Explainer & Metric & $B{=}6$ & $B{=}8$ & $B{=}10$ & $B{=}12$ \\
\midrule
\multicolumn{6}{l}{\emph{MUTAG}} \\
\textsc{GNNExplainer} & Fidelity & 0.218 & 0.256 & 0.285 & 0.322 \\
\textsc{GNNExplainer} & Interpretability & 0.460 & 0.489 & 0.498 & 0.524 \\
\textsc{GNNExplainer} & Stability (VGAE) & 0.665 & 1.248 & 1.701 & 2.571 \\
\textsc{GNNExplainer} & Stability (GNTK) & 1\,280 & 2\,110 & 3\,144 & 5\,647 \\
\textsc{PGExplainer} & Fidelity & 0.234 & 0.256 & 0.278 & 0.331 \\
\textsc{PGExplainer} & Interpretability & 0.965 & 1.039 & 1.169 & 1.494 \\
\textsc{PGExplainer} & Stability (VGAE) & 3.303 & 3.593 & 3.950 & 4.811 \\
\textsc{PGExplainer} & Stability (GNTK) & 1\,155 & 1\,797 & 2\,681 & 4\,706 \\
\midrule
\multicolumn{6}{l}{\emph{BA-2Motif}} \\
\textsc{GNNExplainer} & Fidelity & 0.079 & 0.087 & 0.111 & 0.131 \\
\textsc{GNNExplainer} & Interpretability & 0.026 & 0.027 & 0.029 & 0.031 \\
\textsc{GNNExplainer} & Stability (VGAE) & 1.111 & 1.376 & 1.544 & 1.829 \\
\textsc{GNNExplainer} & Stability (GNTK) & 523.479 & 769.548 & 1\,319 & 2\,478 \\
\textsc{PGExplainer} & Fidelity & 0.178 & 0.240 & 0.280 & 0.312 \\
\textsc{PGExplainer} & Interpretability & 0.025 & 0.026 & 0.026 & 0.027 \\
\textsc{PGExplainer} & Stability (VGAE) & 1.383 & 1.616 & 1.947 & 2.283 \\
\textsc{PGExplainer} & Stability (GNTK) & 377.565 & 690.135 & 959.975 & 1\,957 \\
\bottomrule
\end{tabular}

%% file: variance_table.tex
Every curve in Figures~\ref{fig:sweep-all-metrics} and \ref{fig:tradeoff-vgae} is a single run, and MCTS is stochastic. To quantify the resulting noise floor, we re-ran the proposed method with 3 random seeds on a fixed 20-graph subsample of each dataset, for every explanation budget and under both similarity indices, with the metric-importance controls set to their default values $(0.833, 0.083, 0.083)$. Tables~\ref{tab:var-fidelity}--\ref{tab:var-stability} report the mean and standard deviation across seeds.

 The subsample mean is not the full-dataset mean, so these entries do not reproduce the values plotted in the main text and are not error bars on them; they estimate only the width of the run-to-run distribution. Also, with 3 seeds, the standard deviation is itself imprecise, and should be read as an order of magnitude for the noise rather than a calibrated interval.

\begin{table}[h]
\centering
\small
\caption{Seed-to-seed fidelity spread (mean $\pm$ std over 5 seeds, 20 graphs per run).}
\label{tab:var-fidelity}
\begin{tabular}{llcccc}
\toprule
Dataset & Method & $B=6$ & $B=8$ & $B=10$ & $B=12$ \\
\midrule
MUTAG & Ours (\textsc{VGAE}) & $0.293 \pm 0.006$ & $0.357 \pm 0.010$ & $0.462 \pm 0.017$ & $0.533 \pm 0.015$ \\
MUTAG & Ours (\textsc{GNTK}) & $0.298 \pm 0.005$ & $0.352 \pm 0.005$ & $0.462 \pm 0.010$ & $0.544 \pm 0.014$ \\
\addlinespace
BA-2Motif & Ours (\textsc{VGAE}) & $0.433 \pm 0.027$ & $0.496 \pm 0.041$ & $0.916 \pm 0.061$ & $0.955 \pm 0.039$ \\
BA-2Motif & Ours (\textsc{GNTK}) & $0.461 \pm 0.001$ & $0.523 \pm 0.026$ & $0.935 \pm 0.053$ & $0.973 \pm 0.025$ \\
\addlinespace
BAMultiShapes & Ours (\textsc{VGAE}) & $0.215 \pm 0.004$ & $0.234 \pm 0.005$ & $0.260 \pm 0.008$ & $0.265 \pm 0.007$ \\
BAMultiShapes & Ours (\textsc{GNTK}) & $0.215 \pm 0.003$ & $0.233 \pm 0.006$ & $0.257 \pm 0.010$ & $0.268 \pm 0.007$ \\
\addlinespace
PROTEINS & Ours (\textsc{VGAE}) & $0.227 \pm 0.011$ & $0.255 \pm 0.010$ & $0.261 \pm 0.024$ & $0.263 \pm 0.018$ \\
PROTEINS & Ours (\textsc{GNTK}) & $0.229 \pm 0.015$ & $0.252 \pm 0.012$ & $0.253 \pm 0.020$ & $0.275 \pm 0.007$ \\
\addlinespace
\bottomrule
\end{tabular}
\end{table}

\begin{table}[h]
\centering
\small
\caption{Seed-to-seed interpretability spread (mean $\pm$ std over 5 seeds, 20 graphs per run).}
\label{tab:var-interp}
\begin{tabular}{llcccc}
\toprule
Dataset & Method & $B=6$ & $B=8$ & $B=10$ & $B=12$ \\
\midrule
MUTAG & Ours (\textsc{VGAE}) & $1.193 \pm 0.086$ & $1.414 \pm 0.086$ & $1.472 \pm 0.050$ & $1.469 \pm 0.050$ \\
MUTAG & Ours (\textsc{GNTK}) & $1.181 \pm 0.052$ & $1.337 \pm 0.033$ & $1.419 \pm 0.036$ & $1.382 \pm 0.022$ \\
\addlinespace
BA-2Motif & Ours (\textsc{VGAE}) & $0.073 \pm 0.014$ & $0.178 \pm 0.036$ & $1.927 \pm 0.067$ & $1.949 \pm 0.084$ \\
BA-2Motif & Ours (\textsc{GNTK}) & $0.083 \pm 0.006$ & $0.135 \pm 0.023$ & $1.632 \pm 0.185$ & $1.725 \pm 0.038$ \\
\addlinespace
BAMultiShapes & Ours (\textsc{VGAE}) & $-323.4 \pm 0.1$ & $-323.3 \pm 0.0$ & $-323.2 \pm 0.1$ & $-323.2 \pm 0.1$ \\
BAMultiShapes & Ours (\textsc{GNTK}) & $-323.5 \pm 0.1$ & $-323.3 \pm 0.1$ & $-323.3 \pm 0.1$ & $-323.2 \pm 0.0$ \\
\addlinespace
PROTEINS & Ours (\textsc{VGAE}) & $1.174 \pm 0.126$ & $1.249 \pm 0.120$ & $1.344 \pm 0.062$ & $1.613 \pm 0.113$ \\
PROTEINS & Ours (\textsc{GNTK}) & $1.118 \pm 0.172$ & $1.331 \pm 0.154$ & $1.487 \pm 0.160$ & $1.468 \pm 0.190$ \\
\addlinespace
\bottomrule
\end{tabular}
\end{table}

\begin{table}[h]
\centering
\small
\caption{Seed-to-seed stability spread (mean $\pm$ std over 5 seeds, 20 graphs per run).}
\label{tab:var-stability}
\begin{tabular}{llcccc}
\toprule
Dataset & Method & $B=6$ & $B=8$ & $B=10$ & $B=12$ \\
\midrule
MUTAG & Ours (\textsc{VGAE}) & $29.4 \pm 0.8$ & $35.5 \pm 0.7$ & $40.9 \pm 1.4$ & $46.0 \pm 1.1$ \\
MUTAG & Ours (\textsc{GNTK}) & $14\,898 \pm 241$ & $30\,827 \pm 485$ & $53\,563 \pm 1\,163$ & $80\,839 \pm 2\,670$ \\
\addlinespace
BA-2Motif & Ours (\textsc{VGAE}) & $12.4 \pm 1.3$ & $15.8 \pm 1.4$ & $18.4 \pm 1.8$ & $19.8 \pm 1.7$ \\
BA-2Motif & Ours (\textsc{GNTK}) & $1\,764 \pm 82$ & $2\,735 \pm 716$ & $3\,695 \pm 246$ & $4\,890 \pm 661$ \\
\addlinespace
BAMultiShapes & Ours (\textsc{VGAE}) & $9.387 \pm 0.839$ & $10.4 \pm 0.6$ & $10.9 \pm 1.0$ & $12.1 \pm 0.9$ \\
BAMultiShapes & Ours (\textsc{GNTK}) & $1\,795 \pm 120$ & $2\,774 \pm 116$ & $3\,607 \pm 230$ & $4\,937 \pm 390$ \\
\addlinespace
PROTEINS & Ours (\textsc{VGAE}) & $13.7 \pm 0.8$ & $13.9 \pm 0.2$ & $14.2 \pm 0.4$ & $14.4 \pm 0.7$ \\
PROTEINS & Ours (\textsc{GNTK}) & $1\,346 \pm 86$ & $1\,748 \pm 98$ & $2\,126 \pm 85$ & $2\,513 \pm 245$ \\
\addlinespace
\bottomrule
\end{tabular}
\end{table}


%% file: ba2motif_tradeoff.tex
\begin{figure*}[t]
  \centering
  \includegraphics[width=\textwidth]{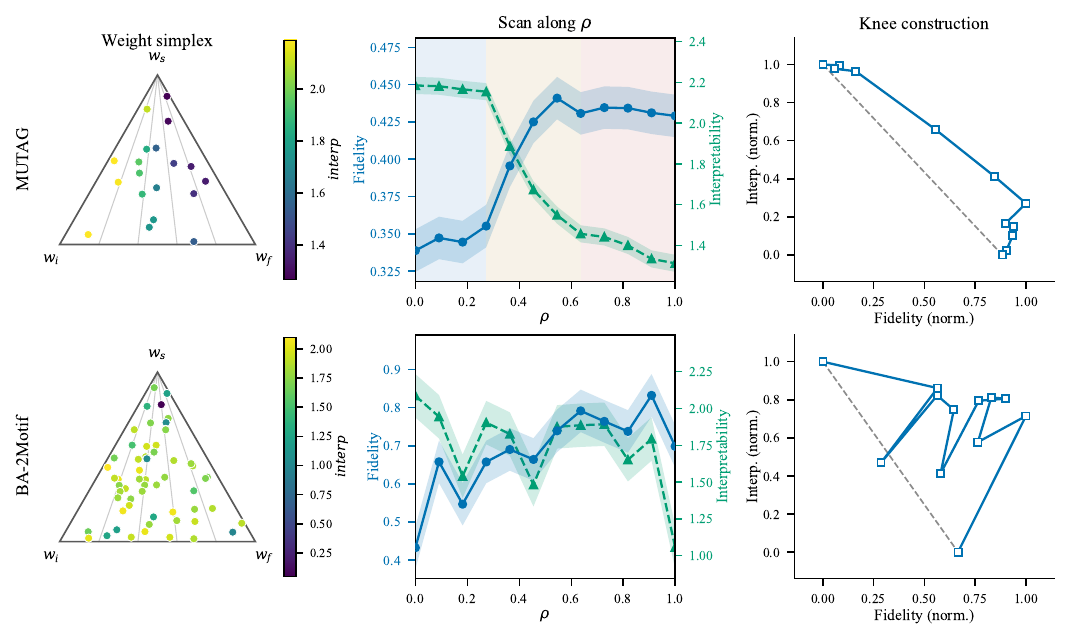}
  \caption{\textbf{The trade-off surface under the GNTK similarity index}, plotted exactly as Figure~\ref{fig:tradeoff-vgae}. The agreement between this figure and Figure~\ref{fig:tradeoff-vgae} rules out an artifact of the similarity index.}
  \label{fig:tradeoff-gntk}
\end{figure*}